\documentclass[10pt,twocolumn,letterpaper]{article}

\usepackage{iccv}
\usepackage{times}
\usepackage{epsfig}
\usepackage{graphicx}
\usepackage{amsmath}
\usepackage{amssymb}

\usepackage{cite}
\usepackage{amsmath,amssymb,amsfonts}
\usepackage{algorithmic}
\usepackage{graphicx}
\usepackage{textcomp}
\usepackage{xcolor}

\usepackage{url}
\usepackage{algorithm,algorithmic}
\usepackage{booktabs,makecell,multirow}

\usepackage[breaklinks=true,bookmarks=false]{hyperref}

\iccvfinalcopy 

\def\iccvPaperID{12} 

\ificcvfinal\fi

\begin{document}

\title{Continuous Emotion Recognition with Audio-visual Leader-follower Attentive Fusion}

\author{Su Zhang\textsuperscript{1,*}, Yi Ding\textsuperscript{1,*}, Ziquan Wei\textsuperscript{2,\thanks{They are equal contributors to this work. Cuntai Guan is the corresponding author. This work is partially supported by the RIE2020 AME Programmatic Fund, Singapore (No. A20G8b0102).}}, Cuntai Guan\textsuperscript{1} \\
\textsuperscript{1}Nanyang Technological University, \textsuperscript{2}Huazhong University of Science and Technology\\
{\tt\small sorazcn@gmail.com, ding.yi@ntu.edu.sg, wzquan142857@hust.edu.cn, ctguan@ntu.edu.sg}
}

\maketitle
\ificcvfinal\thispagestyle{empty}\fi

\begin{abstract}
We propose an audio-visual spatial-temporal deep neural
network with: (1) a visual block containing a pretrained 2D-CNN followed by a temporal convolutional network (TCN); (2) an aural block containing several parallel TCNs; and (3) a leader-follower attentive fusion block combining the audio-visual information. The TCN with large history coverage enables our model to exploit spatial-temporal information within a much larger window length (i.e., 300) than that from the baseline and state-of-the-art methods (i.e., 36 or 48). The fusion block emphasizes the visual modality while exploits the noisy aural modality using the inter-modality attention mechanism. To make full use of the data and alleviate over-fitting, the cross-validation is carried out on the training and validation set. The concordance correlation coefficient (CCC) centering is used to merge the results from each fold. On the test (validation) set of the Aff-Wild2 database, the achieved CCC is $0.463\,(0.469)$ for valence and $0.492\,(0.649)$ for arousal, which significantly outperforms the baseline method with the corresponding CCC of $0.200\,(0.210)$ and $0.190\,(0.230)$ for valence and arousal, respectively. The code is available at https://github.com/sucv/ABAW2.
\end{abstract}

\section{Introduction}
Emotion recognition is the process of identifying human emotion. It plays a crucial role for many human-computer interaction systems. To describe the human state of feeling, psychologists have developed the categorical and the dimensional \cite{sandbach2012static} models. The categorical model is based on several basic emotions. It has been extensively exploited in affective computing largely due to its simplicity and universality. The dimensional model maps the emotion into a continuous space, where the valence and arousal are taken as the axes. It can describe more complex and subtle emotions. This paper focuses on developing a continuous emotion recognition method based on the dimensional model.

Continuous emotion recognition seeks to automatically predict subject's emotional state in a temporally continuous manner. Given the subject's visual, aural, and physiological data which are temporally sequential and synchronous, the system aims to map all the information onto the dimensional space and produces the valence-arousal prediction. The latter is then evaluated against the expert annotated emotional trace using metrics such as the concordance correlation coefficient (CCC). A number of databases, including SEMAINE \cite{mckeown2011semaine}, RECOLA \cite{ringeval2013introducing}, MAHNOC-HCI \cite{soleymani2011multimodal}, SEWA \cite{kossaifi2019sewa}, and MuSe \cite{stappen2021multimodal} have been built for this task. Depending on the subject's context, i.e., controlled or in-the-wild environments, and induced or spontaneous behaviors, the task can be quite challenging due to varied noise levels, illumination, and camera calibration, etc.   Recently, Kollias et al. \cite{zafeiriou2017aff,kollias2019deep,kollias2019face,kollias2019expression,kollias2020analysing,kollias2021affect,kollias2021distribution} build the Aff-Wild2 database, which is by far the largest available in-the-wild database for continuous emotion recognition. The Affective  Behavior  Analysis  in-the-wild  (ABAW) competition \cite{2106.15318} is later hosted using the Aff-Wild2 database.


\begin{figure*}[t]
\centering
\includegraphics[width=\textwidth]{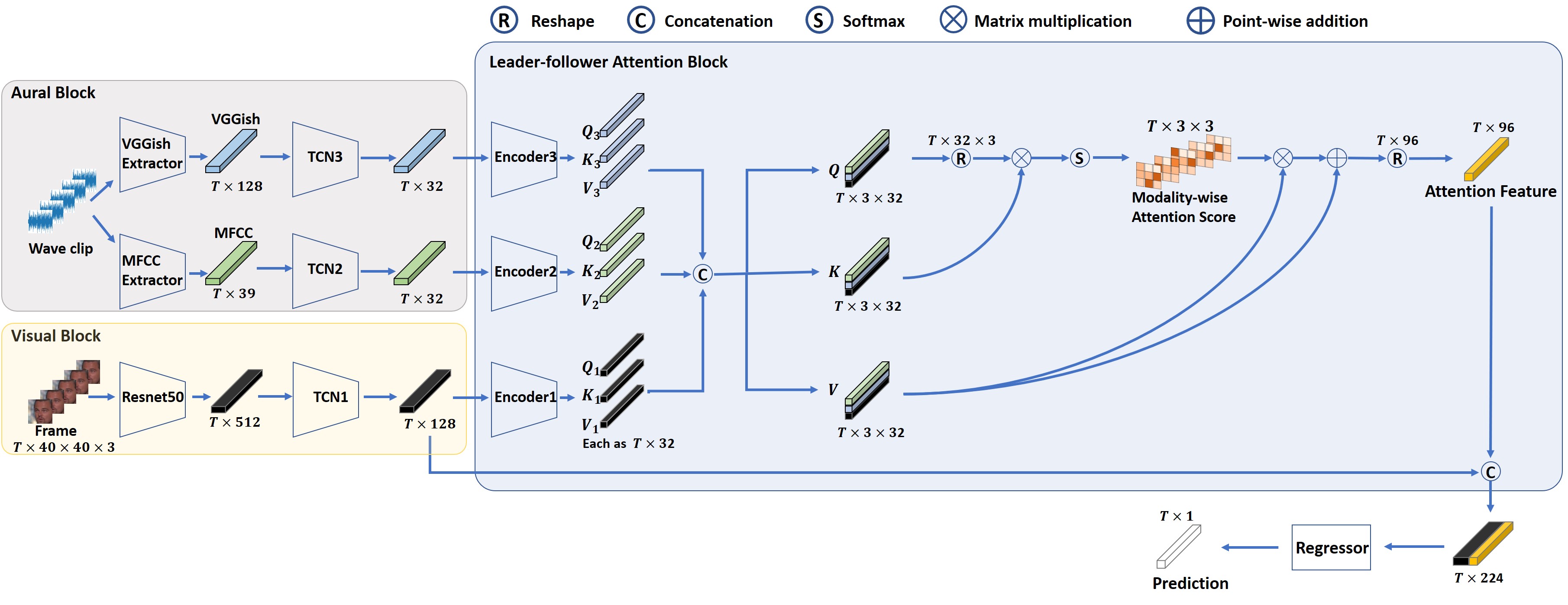}
\caption{The architecture of our audio-visual spatial-temporal model. The model consists of three components, i.e., the visual, aural, and leader-follower attentive fusion blocks. The visual block has a cascade 2DCNN-TCN structure, and the aural block contains two parallel TCN branches. Together with the visual block, the three branches yield three independent spatial-temporal feature vectors. They are then fed to the attentive fusion block. Three independent attention encoders are used. For the $i$-th branch, its encoder consists of three independent linear layers, they adjust the dimension of feature vector producing a query $\mathbf{Q}_i$, a key $\mathbf{K}_i$, and a value $\mathbf{V}_i$. They are then regrouped and concatenated to form the cross-modal counterparts. For example, the cross-modal query $\mathbf{Q}=[\mathbf{Q}_1,\mathbf{Q}_2,\mathbf{Q}_3]$. An attention score is obtained by Eq. \ref{eq:att}. The $3\times 3$ attention score will guide the model to refer to a specific modality at each time step, producing the attention feature. Finally, by concatenating the leading visual features with the attention feature, our model emphasizes the dominant visual modality and makes the inference.}\label{fig:arc}
\end{figure*}

This paper investigates one question, i.e., how to appropriately combine features of different modalities to achieve an "$1+1>2$" performance. Facial expression is one of the most powerful, natural, and universal signals for human beings to convey or regulate emotional states and intentions \cite{darwin2015expression,tian2001recognizing}. And, voice also serves as a key cue for both the emotion production and emotion perception \cite{scherer2007component}. Though we human are good at recognizing emotion from multi-modal information, a straightforward feature concatenation may deteriorate an AI emotion recognition system. In a scene where one subject is watching a video or attending a talk show, more than one voice sources can exist, e.g., from the subject him/herself, the video, the anchor, and the audience.  It is not trivial to design an appropriate fusion scheme so that the subject's visual information and the complementary aural information are captured.


We propose an audio-visual spatial-temporal deep neural network with an attentive feature fusion scheme for continuous valence-arousal emotion recognition. The network consists of three branches, fed by facial images, mfcc, and VGGish \cite{hershey2017cnn} features. A Resnet50 followed by a temporal convolutional network (TCN) \cite{bai2018empirical} is used to extract the spatial-temporal visual feature of the facial images. Two TCNs are employed to extract the spatial-temporal aural feature from the mfcc and VGGish features. The three branches work in parallel and their outputs are sent to the attentive fusion block. To emphasize the dominance of the visual features, which we believe to have the strongest correlation with the label, a leader-follower strategy is designed. The visual feature plays as the leader and owns a skip connection down to the block's output. The mfcc and VGGish features play as the followers, together with the visual feature they are weighted by an attention score. Finally, the leading visual feature and the weighted attention feature are concatenated and a fully-connected layer is used for the regression. To alleviate over-fitting and exploit the available data, a 6-fold cross-validation is carried out on the combined training and validation set of the Aff-Wild2 database. For each emotion dimension, the final inference is determined by the CCC-centering across the 6 trained models \cite{shrout1979intraclass}.

The remainder of the paper is arranged as follows. Section \ref{sec:relate_works} discusses the related works. Section \ref{sec:model_architecture} details the model architecture including the visual, aural, and attentive fusion blocks. Section \ref{sec:implementation_details} elaborates the implementation details including the data pre-processing, training settings, and post-processing. Section \ref{sec:result} provides the continuous emotion recognition results on the Aff-Wild2 database. Section \ref{sec:conclusion} concludes the work.

\section{Related Works}
\label{sec:relate_works}
\subsection{Video-based Emotion Recognition Methods}
Along with the thriving deep CNN-based methods comes two fundamental frameworks of neural networks for video-based emotion recognition. The first type possesses a cascade spatial-temporal architecture. The convolutional neural networks (CNN) are used to extract spatial information, from which the temporal information is obtained by using temporal models such as Time-delay, recurrent neural networks (RNN), or long short-term memory networks (LSTM). The second type combines the two separated steps into one and extracts the spatial-temporal feature using 3D-CNN. Our model belongs to the first type. 

Two issues hinder the performance of the 3D-CNN based emotion recognition methods. First, they have considerably more parameters than 2D-CNN due to extra kernel dimension. Hence, it is more difficult to train. Second, employing 3D-CNN means to preclude the benefits of large-scale 2D facial image databases (such as MS-CELEB-1M \cite{guo2016ms} and VGGFace2 \cite{cao2018vggface2}). 3D-based emotion recognition databases \cite{fanelli20103,yin20063d,zhang2014bp4d} are much fewer. They are mostly based on posed behavior with limited subjects, diversity, and labels.

In this paper, we employ the cascade CNN-TCN architecture. Systematical comparison \cite{bai2018empirical} demonstrated that TCNs convincingly outperform recurrent architectures across a broad range of sequence modeling tasks. With the dilated and casual convolutional kernel and stacked residual blocks, the TCN is capable of looking very far into the past to make a prediction \cite{bai2018empirical}. Compared to many other methods which utilize smaller window length, e.g., with a sequence length of $70$ for AffWildNet \cite{liu2019aff}, or $32$ for the ABAW2020 VA-track champion \cite{deng2020multitask}, ours of length $300$ has achieved promising improvement on the Aff-Wild2 database.

\section{Model Architecture}
\label{sec:model_architecture}
The model architecture is illustrated in Fig. \ref{fig:arc}. In this section, we detail the proposed audio-visual spatial-temporal model and the leader-follower attentive fusion scheme.

\subsection{Visual Block}
The visual block consists of a Resnet50 and a TCN. The resnet50 is pre-trained on the MS-CELEB-1M dataset \cite{guo2016ms} as a facial recognition task, it is then fine-tuned on the FER+ \cite{barsoum2016training} dataset. The Resnet50 is used to extract the independent per-frame features of the video frame sequence, producing the $512$-D spatial encodings. The latter is then stacked and fed to a TCN, generating the $128$-D spatial-temporal visual features. The TCN utilizes $128\times4$ as the $channel\times layer$ setting with a kernel size of $5$ and dropout of $0.1$. Finally, a fully connected layer is employed to map the extracted features onto a $1$-D sequence. Following the labeling scheme of the Aff-Wild2 database where the label frequency equals the video frame rate, each frame of the input video sequence is exactly corresponding to one label point.

\subsection{Aural Block}
The aural block consists of two parallel branches. The $39$-D mfcc and $128$-D VGGish \cite{hershey2017cnn} features are the inputs, respectively. The  mfcc feature is extracted using the OpenSmile Toolkit \cite{eyben2010opensmile}, and the  VGGish feature is obtained from the pre-trained VGGish model \cite{cao2018vggface2}. They are fed to two independent TCNs and yield two $32$-D spatial-temporal aural features. The TCN utilizes $32\times4$ as the $channel\times layer$ setting with the same kernel size and dropout as in the visual block.

\subsection{Leader-follower Attention Block}
The motivation is two-fold. First, we believe that the representability of the multi-modal information is superior to the unimodal counterpart. In addition to the expressive visual information, the voice usually makes us resonate with the emotional context. Second, a direct feature concatenation may deteriorate the performance due to the noisy aural information. Voice separation is still a topic of research. When multiple voice sources exist, it is difficult to obtain the voice components relevant to a specific subject.

The block first maps the feature vectors to query, key, and value vectors by the following procedure. For the $i$-th branch, its encoder consists of three independent linear layers, they adjust the dimension of feature vector producing a query $\mathbf{Q}_i$, a key $\mathbf{K}_i$, and a value $\mathbf{V}_i$. They are then regrouped and concatenated to form the cross-modal counterparts. For example, the cross-modal query $\mathbf{Q}=[\mathbf{Q}_1,\mathbf{Q}_2,\mathbf{Q}_3]$.  After which, the attention feature is calculated as:

\begin{equation}
Attention(\mathbf{Q}, \mathbf{K}, \mathbf{V})=(softmax(\frac{\mathbf{QK}^T}{\sqrt{d_K}})+1)\mathbf{V},
\label{eq:att}
\end{equation}
where $d_K=32$ is the dimension of the key $\mathbf{K}$. After which, the $Attention$ is normalized and concatenated to the leader feature (i.e., the spatial-temporal visual feature in our case) producing the $224$-D leader-follower attention feature. Finally, a fully connected layer is used to yield the inference.

Note that the unimodal version of our model has only the visual block. The inference is obtained based on the $128$-D spatial-temporal visual feature.

\section{Implementation Details}
\label{sec:implementation_details}
\subsection{Database}
Our work is based on Aff-Wild2 database. It consists of 548 videos collected from YouTube. All the video are captured in-the-wild. 545 out of 548 videos contain annotations in terms of valence-arousal. The annotations are provided by four experts using a joystick \cite{cowie2000feeltrace}. The resulted valence and arousal values range continuously in $[-1, 1]$. The final label values are the average of the four raters. The database is split into the training, validation and test sets. The partitioning is done in a subject independent manner, so that every subject's data will present in only one subset. The partitioning produces 346, 68, and 131 videos for the training, validation, and test sets.

\begin{table*}[ht]
\centering
\caption{The validation result in CCC using 6-fold cross validation from the unimodal and multi-moda models. The 6-fold cross-validation is used for data expansion and over-fitting prevention, in which the fold 0 is exactly the original data partitioning provided by ABAW2021.}\label{table:result}
\setlength{\tabcolsep}{4mm}{
\begin{tabular}{ccccccccc}
\toprule
 
\makecell[c]{Emotional\\dimension} &\makecell[c]{Method}&\makecell[c]{ Fold 0}& \makecell[c] {Fold 1}  &\makecell[c]{Fold 2} &\makecell[c]{Fold 3} &\makecell[c]{Fold 4} &\makecell[c]{ Fold 5} &\makecell[c]{Mean}         \\

\midrule
\multirowcell{3}{Valence}&Baseline&$0.210$&$-$&$-$&$-$&$-$&$-$&$-$\\
&Ours-unimodal&$0.425$&$0.614$&$0.437$&$0.523$&$0.526$&$0.516$&$0.507$\\

                   &Ours-multimodal&$0.469$&$0.563$&$0.428$&$0.518$&$0.520$&$0.503$&$0.500$\\
                   
\midrule
\multirowcell{3}{Arousal}&Baseline&$0.230$&$-$&$-$&$-$&$-$&$-$&$-$\\
&Ours-unimodal&$0.647$&$0.651$&$0.548$&$0.573$&$0.587$&$0.584$&$0.598$\\
                   &Ours-multimodal&$0.649$&$0.655$&$0.559$&$0.592$&$0.629$&$0.603$&$0.615$\\

\bottomrule
\end{tabular}}
\end{table*}

\subsection{Preprocessing}
The visual preprocessing is carried out as follows. The cropped-aligned image data provided by the ABAW2021 challenge are used. All the images are resized to $48\times 48\times 3$. Given a trial, the length $N$ is determined by the number of the rows which does not include $-5$. A zero matrix $\mathbf{B}$ of size $N\times 48\times 48\times 3$ is initialized and then iterated over the rows. For the $i$-th row of $\mathbf{B}$, it is assigned as the $i$-th jpg image if it exists, otherwise doing nothing. After which, the matrix $\mathbf{B}$ is saved in npy format and serves as the visual data of this trial.

The aural preprocessing firstly converts all the videos to mono with a $16K$ sampling rate in wav format. The synchronous mfcc and VGGish features are then extracted, respectively. For the mfcc feature, it is extracted using the OpenSmile Toolbox. The settings are the same as provided by AVEC2019 challenge \cite{ringeval2019avec}\footnote{\url{https://github.com/AudioVisualEmotionChallenge/AVEC2019/blob/master/Baseline_features_extraction/Low-Level-Descriptors/extract_audio_LLDs.py}}. Since the window and hop lengths of the short-term Fourier transform are fixed to $25$ms and $10$ms, respectively, the mfcc feature has a fixed frequency of $100$ Hz over all trials. Given the varied labeling frequency and the fixed mfcc frequency, the synchronicity is achieved by pairing the $i$-th label point with the closest-in-time-stamp mfcc feature point. For example, given a label sequence in $30$ Hz, the $\{0, 1, 2, 3\}$-rd label points sampled at $\{0, 0.033, 0.067, 0.100\}$ seconds are paired with the $\{0, 3, 7, 10\}$-th feature points sampled at $\{0, 0.03, 0.07, 0.10\}$ seconds. For all the label rows that do not contain $-5$, the paired mfcc feature points are selected in sequential to form the feature matrix.

For the VGGish feature, it is extracted using the pre-trained VGGish model\footnote{\url{https://github.com/tensorflow/models/tree/master/research/audioset/vggish}}. First, the log-mel spectrogram is extracted and synchronized with the label points using the same operation above. The log-mel spectrogram matrix is then fed into the pre-trained model to extract the synchronized VGGish features. To ensure that the aural features and the labels have the same length, the feature matrices are repeatedly padded using the last feature points. The aural features are finally saved in npy format.

For the valence-arousal labels, all the rows containing $-5$ are excluded. The labels are then saved in npy format.

\subsection{Data Expansion}

The AffWild2 database contains $351$ and $71$ trials in the training and validation sets, respectively. To clarify, a label txt file and its corresponding data are taken as a trial. Note that some videos include two subjects, resulting in two separated cropped-aligned image folders and label txt files, with different suffixes. They are each taken as two trials.

To make full use of the available data and alleviate over-fitting, the cross-validation is employed. By evenly splitting the training set into $5$ folds, we have $6$ folds in total with a roughly equal trial amount, i.e., $70\times 4+71+71$ trials. Note that the $0$-th fold is exactly the original data partitioning. And there is no subject overlap across different folds. The CCC-centering is employed to merge the inference result on the test set.

Moreover, during training and validation, the resampling window has a $33\%$ overlap, resulting in $33\%$ more data. 

\subsection{Training}
Since we employ the 6-fold cross-validation on $2$ emotional dimensions using $2$ models, we have $24$ training instances to run, each takes about $10$Gb VRAM and $1$ to $2$ days. Multiple GPU cards including Nvidia Titan V and Tesla V100 from various servers are used. To maintain the reproducibility, the same Singularity container is shared over the $24$ instances. The code is implemented using Pytorch.

The batch size is $2$. For each batch, the resampling window length and hop length are $300$ and $200$, respectively. I.e., the dataloader loads consecutive $300$ feature points to form a minibatch, with a stride of $200$. For any trials having feature points smaller than the window length, zero padding is employed. For visual data, the random flip, random crop with a size of $40$ are employed for training and only the center crop is employed for validation. The data are then normalized to have $0.5$ mean and standard deviation. For aural data, the data is normalized to have $0$ mean and unit standard deviation.

The CCC loss is used as the loss function. The Adam optimizer with a weight decay of $0.0001$ is employed. The learning rate (LR) and minimal learning rate (MLR) are set to $1e-5$ and $1e-6$, respectively. The \textit{ReduceLROnPlateau} scheduler with a patience of $5$ and factor of $0.1$ is employed based on the validation CCC. The maximal epoch number and early stopping counter are set to $100$ and $20$, respectively. Two groups of layers for the Resnet50 backbone are manually selected for further fine-tuning, which corresponds to the whole layer4 and the last three blocks of layer3. 

The training strategy is as follows. The Resnet50 backbone is initially fixed except the output layer. For each epoch, the training and validation CCC are obtained. If a higher validation CCC appears, then reset the counter to zero, otherwise increase the counter by one. If the counter reaches \textit{Patience}, then reduce $LR$ to $MLR$. Next time when the counter reaches $\textit{Patience}$, release one group of backbone layers (started from the layer4 group) for updating and reset the counter. At the end of each epoch, the current best model state dictionary is loaded. The training is stopped if i) there is no remaining backbone layer group to release,  ii) the counter reaches the \textit{Early Stopping Counter}, or iii) the epoch reaches the \textit{Maximal Epoch}. Note that the valence and arousal models are trained separately.

\begin{figure}[!htbp]
\centering
\includegraphics[width=.5\textwidth]{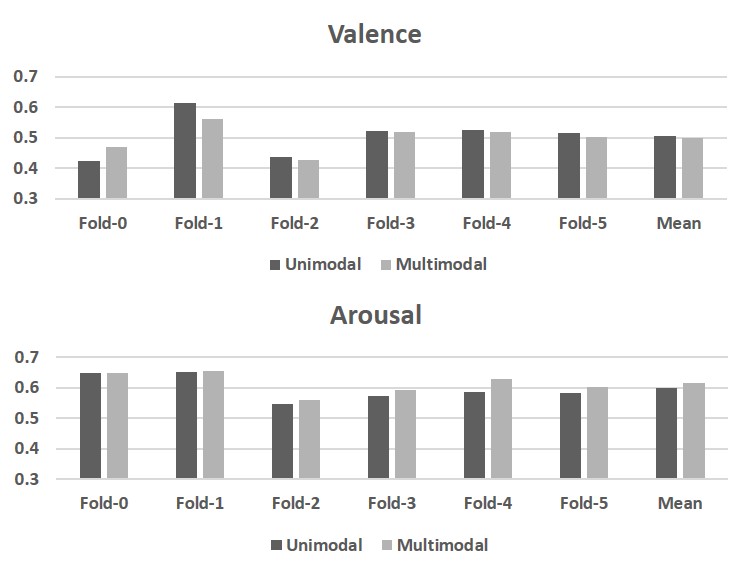}
\caption{The 6-fold validation result in CCC obtained by our unimodal and multimodal models. Note that the fold 0 is exactly the original data partitioning provided by ABAW2021.}\label{fig:bar}
\end{figure}

\subsection{Post-processing}

The post-processing consists of CCC-centering and clipping. 
Given the predictions from 6-fold cross-validation, the CCC-centering aims to yield the weighted prediction based on the inter-class correlation coefficient (ICC) \cite{ringeval2019avec}. This technique has been widely used in many emotion recognition challenges \cite{valstar2016avec,ringeval2017avec,ringeval2018avec,ringeval2019avec} to obtain the gold-standard labels from multiple raters, by which the bias and inconsistency among individual raters are compensated. The clipping ensures that the inference is truncated within the interval $[-1, 1]$, i,e., any values larger or smaller than $1$ or $-1$ are set to $1$ or $-1$. respectively. 

In this work, two strategies, i.e., clipping-then-CCC-centering and CCC-centering-then-clipping are utilized. They are called early-clipping and late-clipping.

\section{Result}
\label{sec:result}

\subsection{Validation Result}
The validation results of the 6-fold cross-validation on valence and arousal are reported in Table \ref{table:result}. Two types of models are trained. The unimodal model is fed by video frame only, and the multimodal model is fed by video frame, mfcc, and VGGish features. 

For fold $0$, namely the original data partitioning, both of our unimodal  and multimodal models have significantly outperformed the baseline. For other five folds, interestingly, the multimodal information has positive and negative effects on the arousal and valence dimensions, respectively, as shown in Figure \ref{fig:bar}. We therefore hypothesize that the annotation protocol weighs more on aural perspective for the arousal dimension.

\subsection{Test Result}
The comparison of our model against the baseline and state-of-the-art methods on the test set are shown in Table \ref{table:overall2}.

\begin{table}[ht]
\centering
\caption{The overall test results in CCC. UM and MM denote our unimodal and multimodal models, respectively. CV denotes cross-validation. EC and LC denote early-clipping and late-clipping. The bold fonts indicate the best result from ours and other teams. }\label{table:overall2}
\vspace{0.25cm}
\begin{tabular}{|p{3cm}|l|l|l|}
\hline
Method                     & Valence & Arousal & Mean  \\ \hline
Baseline                   & 0.200             & 0.190             & 0.195  \\ \hline
ICT-VIPL-VA \cite{zhang2020m} & 0.361 & 0.408 & 0.385\\ \hline
NISL2020 \cite{deng2020multitask} & 0.440 & 0.454 & 0.447 \\ \hline
NISL2021 \cite{deng2021towards} & \textbf{0.533} & \textbf{0.454} & \textbf{0.494} \\ \hline
Netease Fuxi Virtual Human \cite{zhang2021prior} & \multirow{2}{*}{0.486} & \multirow{2}{*}{0.495} & \multirow{2}{*}{0.491} \\\hline
Morphoboid \cite{vu2021multitask} & 0.505 & 0.475 & 0.490 \\ \hline
STAR \cite{wang2021multi} & 0.478 & 0.498 & 0.488 \\ \hline
UM    & 0.267            & 0.303            & 0.285 \\ \hline
UM-CV-EC   & 0.264            & 0.276            & 0.270 \\ \hline
UM-CV-LC   & 0.265            & 0.276            & 0.271 \\ \hline
MM  & 0.455            & 0.480            & 0.468 \\ \hline
MM-CV-EC   & 0.462            & 0.492            & 0.477 \\ \hline
MM-CV-LC   & \textbf{0.463}            & \textbf{0.492}            & \textbf{0.478} \\ \hline
\end{tabular}
\end{table}

First and foremost, the multimodal model achieves great improvement against the unimodal counterpart, which is up to $70.41\%$ and $58.42\%$ gain (by comparing UM against MM in table \ref{table:overall2}) over the valence and arousal, respectively. The employment of cross-validation provides incremental improvement on the multimodal result. 

We can also see that the three unimodal scenarios and multimodal scenarios have a sharp performance gap, whereas on the validation set the gap is incremental. We hypothesize that the unimodal models, fed only by visual information, suffer from over-fitting and insufficient robustness on the test set. The issue is alleviated by the fusion with aural information. Further investigation will be carried out in our future work using other audio-visual databases where labels of the test set are available.

\section{Conclusion}
We proposed an audio-visual spatial-temporal deep neural network with an attentive feature fusion scheme for continuous valence-arousal emotion recognition. The model consists of a visual block, an aural block, and a leader-follower attentive fusion block. The latter achieves the cross-modality fusion by emphasizing the leading visual modality while exploiting the noisy aural modality. Experiments are conducted on the Aff-Wild2 database and promising results are achieved. The achieved CCC on test (validation) set is $0.463\,(0.469)$ for valence and $0.492\,(0.649)$ for arousal, which significantly outperforms the baseline method with the corresponding CCC of $0.200\,(0.210)$ and $0.190\,(0.230)$ for valence and arousal, respectively.

\label{sec:conclusion}
\bibliographystyle{IEEEtran}
\bibliography{ref}

\end{document}